\documentclass[lettersize,journal]{IEEEtran}

\usepackage[T1]{fontenc}    
\usepackage[hidelinks]{hyperref}       
\usepackage{url}            
\usepackage{booktabs}       
\usepackage{amsfonts}       
\usepackage{nicefrac}       
\usepackage{microtype}      
\usepackage[table]{xcolor}         
\usepackage{amsmath,amssymb}
\usepackage{graphicx}
\usepackage[capitalise,noabbrev]{cleveref}
\usepackage{multirow}
\usepackage{colortbl}
\definecolor{LightGray}{gray}{0.9}
\usepackage{float}
\usepackage{cite}
\usepackage{array}

\title{From Alignment to Fusion in 3D Vision-Language}

\author{Xueqi Qiu, Xingyu Miao, Jingjing~Deng, Haoran~Duan,  Yang~Long, and Ling~Shao,~\IEEEmembership{Fellow,~IEEE}%
\thanks{Xueqi Qiu and Xingyu Miao contributed equally to this work.}%
\thanks{Xueqi Qiu, Xingyu Miao and Yang Long are with the Department of Computer Science, Durham University, UK (e-mail: xueqi.qiu@durham.ac.uk; xingyu.miao@durham.ac.uk;  yang.long@durham.ac.uk).}%
\thanks{Jingjing Deng is with the School of Engineering Mathematics and Technology, University of Bristol, Bristol, UK (e-mail: jingjing.deng@bristol.ac.uk).}%
\thanks{Haoran Duan is with the Department of Automation, Tsinghua University, Beijing, China (e-mail: haoran.duan@ieee.org).}%
\thanks{Ling Shao is with the UCAS-Terminus AI Lab, University of Chinese Academy of Sciences, Beijing 100049, China (e-mail: ling.shao@ieee.org).}%
\thanks{Yang Long and Haoran Duan are the corresponding authors.}}

\begin{document}

\maketitle

\begin{abstract}
Unified 3D vision-language systems must combine complementary geometry, scale, and appearance cues while supporting tasks from instance segmentation to language-guided reasoning. Existing methods often process point clouds, voxel grids, and multi-view images independently; directly combining these heterogeneous representations may leave substantial feature discrepancy unresolved, while subsequent unconstrained adaptation may distort their internal geometry. We propose an ``\textit{align-then-fuse}" framework that first applies triple pairwise cosine alignment to establish segment-level correspondence across the three representations and then retrieves task-conditioned features with a prompt-guided query decoder. Before fusion, representation-specific query features are transformed by learnable mappings constrained to the special orthogonal group. These mappings preserve inner products and Euclidean distances within each representation, permitting controlled representation-specific re-parameterisation without arbitrarily distorting its internal geometry. The transformed features are subsequently combined through Adaptive Fusion under downstream task supervision. Experiments cover eight datasets for instance segmentation, visual grounding, question answering, and dense captioning. Compared with PQ3D, the model improves average precision by 3.2 points on ScanNet200 and grounding accuracy by 2.9, 10.6, 4.6, and 4.1 points on ScanRefer, Nr3D, Sr3D, and Multi3DRefer, respectively, while also improving performance on ScanQA, SQA3D, and Scan2Cap. Ablations further support the complementary roles of alignment and orthogonal re-parameterisation and the effectiveness of Adaptive Fusion.
\end{abstract}


\section{Introduction}
In real-world scenarios, achieving a joint understanding of 3D environments and natural language is essential for constructing agents with practical operational abilities \cite{saycan, sayplan, embodied, huang2023embodied, embodiedlang, teach, 3d-vista}. In recent years, with the continuous development of 3D vision-language tasks and their related datasets, significant progress has been made in areas ranging from fundamental 3D semantic and instance segmentation \cite{scannet200, replica} to visual grounding \cite{scanents, referit3d, scanrefer, jia2024sceneverse, multi3drefer}, 3D question answering \cite{scanqa, sqa3d, 3dgqa}, 3D dense captioning \cite{scan2cap}, and open-vocabulary 3D understanding \cite{guo2026semantic, semab, kerr2023lerf, takmaz2023openmask3d}.

Previous research often employed task-specific models built on a single type of 3D representation \cite{vil3dref,scanqa,viewrefer,transrefer3d,3dsps,schult2023mask3d}, while more recent studies increasingly incorporate multiple scene representations within shared 3D vision-language frameworks \cite{3d-vista,zhu2024unifying,ll3da}. Earlier multimodal 3D methods demonstrated the benefit of combining image and 3D information through spatial projection, feature augmentation, or cross-modal interaction \cite{liang2018deep,liang2019multi,vora2020pointpainting,wang2021pointaugmenting,bai2022transfusion,chen2022deformable}. More recent approaches address representation heterogeneity more explicitly. Inst3D-LMM integrates multi-view semantics with matched 3D instances \cite{yu2025inst3d}, while DSPNet selects question-relevant views before combining image and point-cloud features \cite{luo2025dspnet}. CrossOver maps heterogeneous scene modalities into a shared embedding space \cite{sarkar2025crossover}, whereas CUA-O3D and Proxy3D reduce feature heterogeneity through uncertainty-aware knowledge aggregation \cite{li2025cross} and semantic-aware proxy construction \cite{jiang2026proxy3d}, respectively.
These developments establish increasingly effective connections among heterogeneous scene representations, but the underlying features are still generated by representation-specific encoding processes before being jointly used for prediction. Consequently, incorporating these heterogeneous representations into a shared prediction architecture does not by itself ensure their latent compatibility before joint prediction.

\begin{figure*}[]
    \centering
    \includegraphics[width=1\textwidth]{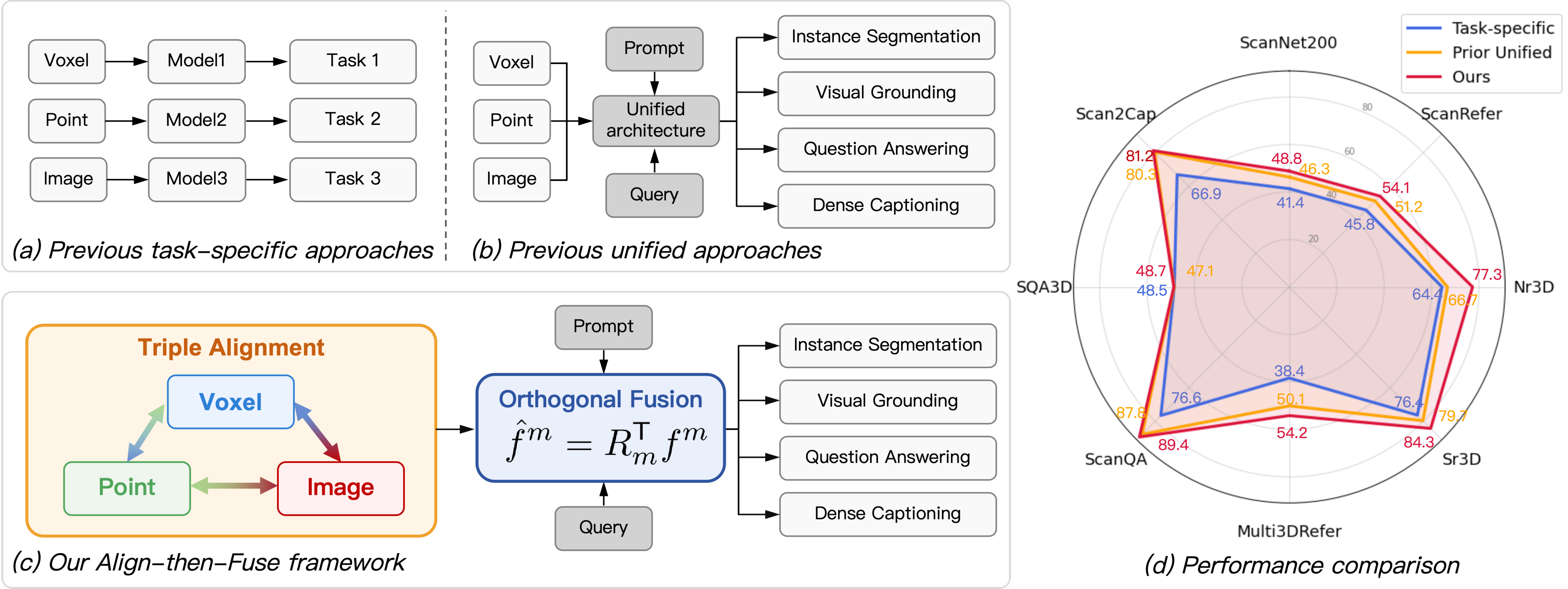}
\caption{\textbf{Overview and performance of our unified 3D vision-language framework.} 
We illustrate the progression from previous task-specific and unified approaches in (a) and (b) to our align-then-fuse framework in (c), which integrates voxel, point-cloud, and image representations through Triple Alignment and Orthogonal Fusion. The performance comparison across eight benchmarks is shown in (d).}
    \label{main}
\end{figure*}

To address this issue, we propose an ``\textit{align-then-fuse}'' framework for multi-representation 3D vision-language understanding. Because features produced by different encoders can remain misaligned before joint prediction, we first introduce Triple Alignment to enforce pairwise cosine correspondence among spatially matched point-cloud, voxel, and multi-view features before prompt-guided decoding. Alignment improves cross-representation correspondence, while the aligned features still retain representation-specific parameterizations before fusion. However, unconstrained transformations may arbitrarily alter the geometry already encoded within each feature space. We therefore constrain each representation-specific mapping to the special orthogonal group $\mathrm{SO}(D)$. The resulting transformations preserve pairwise Euclidean distances and inner products within each representation while permitting learnable re-parameterisation. Finally, the transformed features are combined through Adaptive Fusion and jointly optimized under downstream task supervision. Experiments on various benchmarks spanning instance segmentation, visual grounding, question answering, and dense captioning demonstrate broad improvements over the displayed baselines and support the effectiveness of explicitly aligning heterogeneous representations before their controlled adaptation and fusion.

In summary, our contributions are fourfold:

\begin{itemize}
    \item We investigate two closely related challenges in unified 3D vision-language learning: representation heterogeneity before joint modeling and latent-basis discrepancy during cross-representation fusion.
    \item To address representation heterogeneity, we introduce Triple Alignment to establish explicit pairwise correspondence among spatially matched point-cloud, voxel, and multi-view features.

\item To address latent-basis discrepancy during fusion, we introduce representation-specific orthogonal re-parameterisation based on $\mathrm{SO}(D)$, allowing learnable adaptation while preserving intra-representation geometry, followed by Adaptive Fusion.
    \item We evaluate our approach on eight 3D vision--language datasets: ScanNet200 \cite{scannet200} for instance segmentation; ScanRefer \cite{scanrefer}, ReferIt3D (Nr3D and Sr3D) \cite{referit3d}, and Multi3DRefer \cite{multi3drefer} for visual localization; ScanQA \cite{scanqa} and SQA3D \cite{sqa3d} for question answering; and Scan2Cap \cite{scan2cap} for dense description. In extensive experiments, we demonstrate that our method outperforms the baseline.
\end{itemize}

\begin{figure*}
    \centering
    \includegraphics[width=1\linewidth]{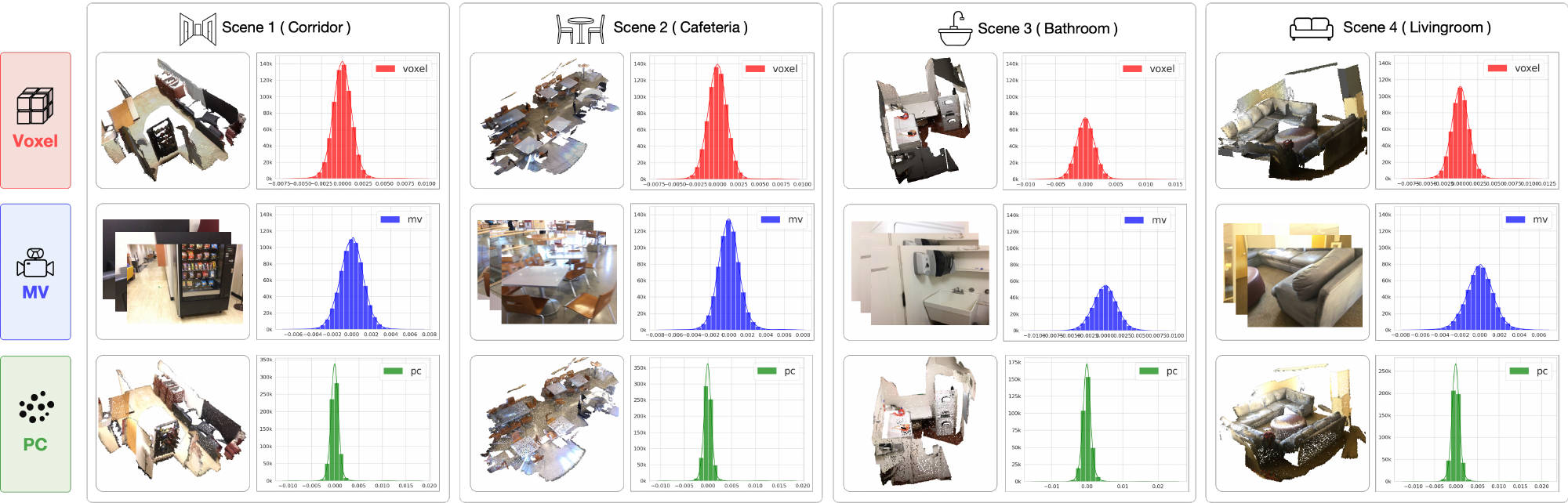}
    \caption{\textbf{Feature distribution discrepancy across heterogeneous 3D representations.}
Feature distributions of voxel, multi-view (MV), and point-cloud (PC) representations are visualized for four representative scenes. Despite describing the same physical scenes, the three representations exhibit distinct distributional characteristics, illustrating the representation discrepancy induced by heterogeneous encoding processes.}
    \label{scene}
\end{figure*}

\section{Related Work}

\subsection{Query-Based 3D Vision-Language Learning}

3D vision-language learning has expanded from individual perception tasks to a broad range of language-conditioned scene understanding problems, including segmentation \cite{scannet200,lee2025mosaic3d}, visual grounding \cite{scanents,referit3d,vil3dref,scanrefer,multi3drefer}, question answering \cite{scanqa,sqa3d}, and dense captioning \cite{scan2cap}. Early methods were predominantly developed for individual tasks, with task-specific architectures and prediction heads tailored to different forms of 3D perception and reasoning.

Query-based prediction provides a natural route toward a more unified formulation. Originating from set-based object prediction \cite{dong2021solq,inst-as-queries,fastinst,hu2021istr,zhu2021deformable}, query representations have been extended to mask prediction \cite{mask2former}, language-guided segmentation \cite{ding2021vision,language-as-queries}, and multimodal prediction \cite{perceiver-io,blip,li2023blip,eq-paradigm,xdecoder}. Rather than defining a separate prediction mechanism for each task, latent queries interact with visual representations and can be conditioned by different forms of prompts, providing a shared interface between input guidance and downstream outputs.

This paradigm has subsequently been adopted in 3D scene understanding. OpenMask3D \cite{takmaz2023openmask3d}, OpenScene \cite{peng2023openscene}, PLA \cite{ding2023pla}, and Mosaic3D \cite{lee2025mosaic3d} associate 3D regions or proposals with vision-language representations for open-vocabulary recognition and segmentation. More recent methods extend query-based interaction beyond category-level prediction toward language-conditioned scene understanding. LL3DA \cite{ll3da} introduces language-guided interaction with 3D scene features, while PQ3D \cite{zhu2024unifying} employs a shared set of promptable queries to accommodate multiple 3D vision-language tasks within a common prediction framework. These developments mark a transition from task-specific 3D vision-language pipelines toward query-based formulations in which diverse prediction objectives can be expressed through a shared interaction mechanism.

\subsection{3D Multimodal Fusion}

Point clouds, voxel grids, and multi-view images provide complementary descriptions of a 3D scene. Point clouds preserve detailed geometric observations, voxel grids impose regular spatial structure, and multi-view images provide dense appearance information. These representations, however, arise from different sampling processes and are typically processed by independently designed encoders, resulting in representation-specific feature spaces with no inherent guarantee of latent compatibility.

Earlier multimodal 3D methods project image features into a shared spatial representation for intermediate fusion \cite{liang2018deep,liang2019multi}, or augment point features with image-derived semantic and appearance cues \cite{vora2020pointpainting,wang2021pointaugmenting}. Subsequent approaches introduce cross-modal attention and deformable interaction to support more flexible information exchange between image and 3D representations \cite{bai2022transfusion,chen2022deformable,yang2022deepinteraction,yan2023cross,xie2023sparsefusion}. These methods demonstrate the effectiveness of combining complementary representations, although their interaction mechanisms are typically developed around a particular representation pair or downstream setting.

More recent studies address representation heterogeneity more explicitly. Inst3D-LMM \cite{yu2025inst3d} injects multi-view semantics into matched 3D instances and models their spatial relations, whereas DSPNet \cite{luo2025dspnet} selects question-relevant views before integrating image and point-cloud features. CrossOver \cite{sarkar2025crossover} maps multiple, potentially incomplete scene modalities into a shared embedding space, while CUA-O3D \cite{li2025cross} and Proxy3D \cite{jiang2026proxy3d} reduce feature heterogeneity through uncertainty-aware knowledge aggregation and semantic-aware proxy construction, respectively. These approaches provide effective mechanisms for connecting heterogeneous scene observations, but the resulting features are still produced by representation-specific encoding processes before being combined for prediction. Consequently, their latent compatibility is not guaranteed by the shared prediction architecture itself. This motivates an explicit treatment of feature compatibility before joint prediction, together with a controlled mechanism for adapting and combining heterogeneous representations without unnecessarily altering the structure already encoded within each feature space.

\begin{figure*}[t]
    \centering
    \includegraphics[width=1\textwidth]{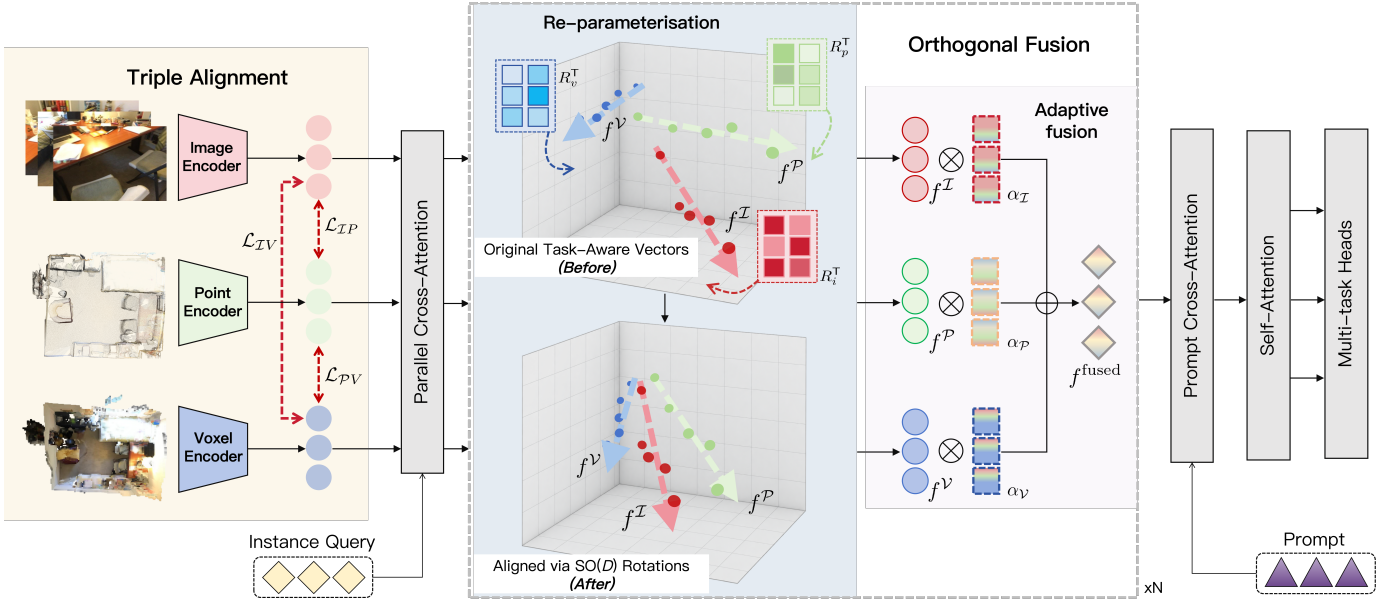}
    \vspace{-1em}
    \caption{\textbf{Architecture of the proposed ``\textit{align-then-fuse}'' framework.}
Point-cloud, voxel, and multi-view image features are first coordinated through Triple Alignment to establish cross-representation correspondence. The resulting representation-specific features are then adapted by orthogonal re-parameterisation, which preserves their intra-representation geometry, and combined through Adaptive Fusion. The fused representation is subsequently conditioned on the task prompt and processed by the multi-task prediction heads for downstream 3D vision-language tasks.}
    \label{pipeline}
    
\end{figure*}

\section{Preliminaries}

\textbf{Task Prompt Encoding.}
We formulate different 3D vision-language tasks using unified task prompts of two types: textual prompts, including object categories, referring expressions, and questions; and numerical prompts, including 3D bounding boxes and 3D locations. Textual prompts are encoded using a pretrained CLIP model \cite{clip}, while numerical prompts are projected to the same hidden dimension through learnable fully connected layers. The resulting prompt features provide task-specific conditioning for downstream prediction.

\textbf{Multi-Representation 3D Encoding.}
To encode a 3D scene, three representations are considered: point clouds, voxel grids, and multi-view images.  For point clouds, the entire set of 3D points is partitioned into segments. In each segment, 1,024 points are sampled and normalized to a unit sphere. These points are processed through a pretrained PointNet++ backbone \cite{pointnet, pointnet++} to obtain point features $\mathcal P = \{p_1, p_2, \dots, p_M\} \in \mathbb{R}^{M \times D}$, where $M$ denotes the number of segments. Voxel features are extracted by discretizing the 3D scene into voxels. The voxel grid is processed with a sparse convolutional U-Net that includes both downsampling and upsampling stages to capture hierarchical details. Features from each voxel are pooled to their corresponding segments and projected into the hidden dimension to form $\mathcal V = \{v_1, v_2, \dots, v_M\} \in \mathbb{R}^{M \times D}$. Multi-view image features are derived by computing per-pixel embeddings using a pretrained segmentation model (e.g., OpenSeg \cite{openseg}). The 2D pixels are back-projected into the 3D space, and features from multiple views are averaged to form image features $\mathcal I = \{i_1, i_2, \dots, i_M\} \in \mathbb{R}^{M \times D}$.

\textbf{Prompt-Guided Query Decoding (Multi-Task Heads).}
The fused scene representation and task prompt are processed by a Transformer-style query decoder with $N_q$ learnable instance queries,
\begin{equation}
Q_l\in\mathbb{R}^{N_q\times D},
\end{equation}
whose positional embeddings are initialized from farthest-point-sampled 3D anchors \cite{fourier-pos}. At each decoder layer, the queries aggregate scene information through cross-attention, interact with the task prompt for task conditioning, and are further refined by spatial self-attention. The resulting prompt-conditioned queries are finally fed to task-specific prediction heads.

\section{Method}

\subsection{Problem Formulation}

\begin{itemize}
    \item \textbf{Modal heterogeneity.} Prior work \cite{zhu2024unifying,ll3da} employs prompt-based queries to extract features from point clouds, multi-scale voxels, and multi-view images. These representations are typically processed in parallel and are either selected according to downstream tasks or mapped into a shared feature space. However, as shown in \Cref{scene}, different representations exhibit distinct feature distributions even for the same scene, revealing a persistent discrepancy among heterogeneous representations. This motivates explicit feature alignment before their joint use.
    \item \textbf{Latent-basis discrepancy.} Alignment alone does not require independently encoded representations to share the same latent parameterization. After alignment, representation-specific adaptation can therefore still be beneficial to retain the flexibility of individual representations before fusion. We refer to this remaining mismatch between independently parameterized feature spaces as latent-basis discrepancy. Applying unconstrained transformations at this stage, however, may arbitrarily distort their encoded feature geometry. We therefore seek a constrained re-parameterisation that can adapt each representation-specific feature space while preserving its internal geometric structure.
\end{itemize}

To address these challenges, we adopt an ``\textit{align-then-fuse}'' strategy. We first explicitly align the features of point-cloud, voxel, and multi-view image representations. The resulting features are then adapted through representation-specific orthogonal re-parameterisation, which permits learnable transformation while preserving pairwise geometry within each representation. The adapted features are subsequently combined through Adaptive Fusion, yielding a fused scene representation that is passed to the multi-task prediction heads for downstream prediction. The overall formulation therefore separates cross-representation correspondence from representation-specific adaptation, allowing heterogeneous 3D features to be coordinated without unconstrained distortion before fusion.

\subsection{Alignment for Representation Heterogeneity}

Features extracted from multi-view images, point clouds, and voxel grids are generated by heterogeneous encoders and may exhibit substantial representation discrepancy. We therefore introduce Triple Alignment, a triple pairwise cosine alignment objective that explicitly encourages corresponding scene-segment features to occupy compatible directions before task-conditioned decoding.

Given multi-view image features $\mathcal{I}\in\mathbb{R}^{M\times D}$, point-cloud features $\mathcal{P}\in\mathbb{R}^{M\times D}$, and voxel features $\mathcal{V}\in\mathbb{R}^{M\times D}$, we formulate Triple Alignment as a segment-level triple pairwise cosine alignment objective. We first define the cosine distance between two feature vectors $x,y\in\mathbb{R}^{D}$ as
\begin{equation}
\mathcal{L}(x,y)
=
1-\frac{x^{\mathsf T}y}{\|x\|_{2}\|y\|_{2}},
\end{equation}
where $x^{\mathsf T}y$ denotes the inner product and $\|\cdot\|_{2}$ denotes the Euclidean norm.

For the $m$-th scene segment, let $i_m$, $p_m$, and $v_m$ denote the corresponding image, point-cloud, and voxel features, respectively. Triple Alignment explicitly considers all three unordered representation pairs:
\begin{equation}
\begin{aligned}
\mathcal{L}_{\mathcal{I}\mathcal{P}}
&=
\frac{1}{M}\sum_{m=1}^{M}\mathcal{L}(i_m,p_m),\\
\mathcal{L}_{\mathcal{I}\mathcal{V}}
&=
\frac{1}{M}\sum_{m=1}^{M}\mathcal{L}(i_m,v_m),\\
\mathcal{L}_{\mathcal{P}\mathcal{V}}
&=
\frac{1}{M}\sum_{m=1}^{M}\mathcal{L}(p_m,v_m).
\end{aligned}
\end{equation}

The overall Triple Alignment objective is defined as
\begin{equation}
\mathcal{L}_{\mathrm{align}}
=
w_{\mathcal{I}\mathcal{P}}\mathcal{L}_{\mathcal{I}\mathcal{P}}
+
w_{\mathcal{I}\mathcal{V}}\mathcal{L}_{\mathcal{I}\mathcal{V}}
+
w_{\mathcal{P}\mathcal{V}}\mathcal{L}_{\mathcal{P}\mathcal{V}},
\label{eq:align}
\end{equation}
where $w_{\mathcal{I}\mathcal{P}}$, $w_{\mathcal{I}\mathcal{V}}$, and $w_{\mathcal{P}\mathcal{V}}$ control the contributions of the image--point, image--voxel, and point--voxel alignment terms, respectively. Minimizing \Cref{eq:align} encourages corresponding segment features from the three representations to occupy compatible directions in the latent space before task-conditioned decoding. Triple Alignment thereby establishes explicit pre-fusion correspondence and improves cross-representation comparability without requiring the three representation distributions to become identical.

\begin{figure*}[t]
    \centering
    \includegraphics[width=1\linewidth]{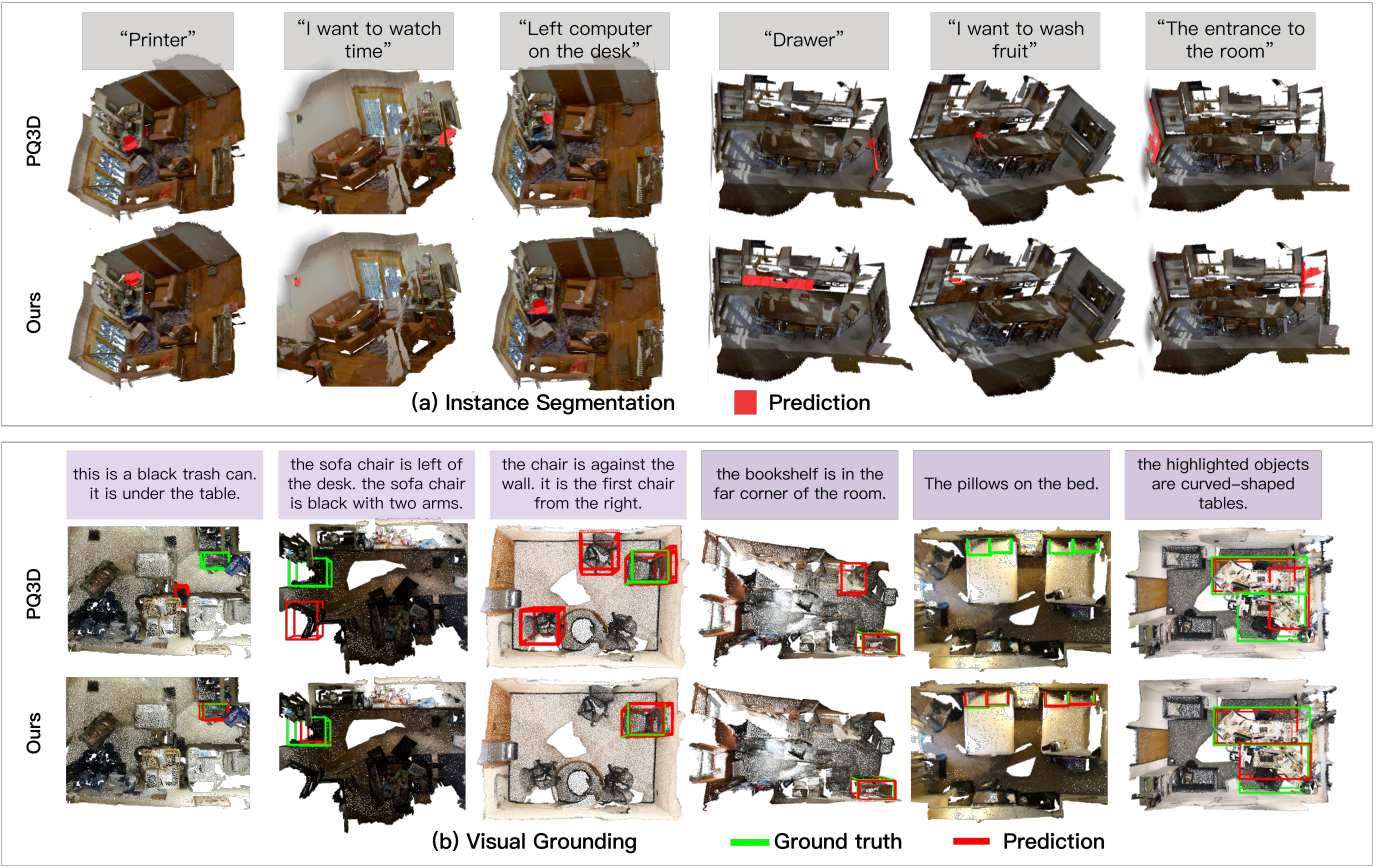}
    \caption{\textbf{Qualitative results on (a) instance segmentation and (b) visual grounding.}
The results demonstrate accurate instance-level scene understanding and reliable localization of language-referred objects across diverse 3D scenes.}
    \label{fig_qua_1}
\end{figure*}

\subsection{Orthogonal Fusion with Re-parameterisation}

Although Triple Alignment improves correspondence among the three representations, the queried features are still generated through independently parameterized representation streams. Directly applying unconstrained linear adapters before fusion may introduce arbitrary scaling, shearing, and anisotropic distortion, thereby altering the internal feature geometry of each stream. We therefore restrict the modality-specific adapters to the special orthogonal group $\mathrm{SO}(D)$. This design does not assume that the residual cross-representation discrepancy is generated by a rotation. Instead, orthogonality serves as a geometry-preserving constraint that allows each representation to be adaptively re-parameterized without changing its internal pairwise structure.

Given the aligned scene representations, each query independently retrieves features from the image, point-cloud, and voxel streams through parallel cross-attention. This produces three query-conditioned vectors, $f^{\mathcal I}, f^{\mathcal P}, f^{\mathcal V}\in\mathbb{R}^{D}$, which retain their representation-specific parameterizations before cross-representation fusion. We apply orthogonal re-parameterisation to the three query-conditioned features and subsequently combine them through adaptive fusion. The fused query representation is then passed to the subsequent query decoding and task-specific prediction heads.

\paragraph{Group preliminaries}
The set of real matrices satisfying $M^{\mathsf T}M=I$ forms the orthogonal group $\mathrm{O}(D)$. Its subset with $\det M=1$ is the special orthogonal group $\mathrm{SO}(D)$, whose elements are pure rotations. For any $R\in\mathrm{SO}(D)$ and any $x,y\in\mathbb{R}^{D}$, the action of $R$ preserves Euclidean distance and inner product:
\begin{equation}
\|Rx - Ry\|_{2} = \|x - y\|_{2}
\quad\text{and}\quad
\langle Rx, Ry\rangle = \langle x,y\rangle.
\label{eq:isometry}
\end{equation}

\paragraph{Why QR decomposition}
To obtain a differentiable orthogonality-constrained adapter, we project an unconstrained square matrix onto an element of $\mathrm{SO}(D)$. The thin QR decomposition $ A = QR $
meets this requirement because its orthogonal factor $Q$ already lies in $\mathrm{O}(D)$ and can be converted to a proper rotation with at most one column-sign flip.
We adopt QR decomposition as a differentiable and computationally convenient way to obtain the orthogonal factor \cite{roberts2026qr}, with efficient implementations available in modern deep-learning frameworks.
Most importantly, the orthogonality of $Q$ preserves Euclidean distances and inner products within each representation stream, preventing the adapter from arbitrarily distorting its internal geometry before fusion.

\paragraph{Isometric re-parameterisation}
For each modality $m\in{\mathcal I,\mathcal P,\mathcal V}$ we keep a learnable square matrix $A_{m}\in\mathbb{R}^{D\times D}$ and compute its thin QR decomposition $A_{m}=O_{m}U_{m}$ at every forward pass, where $O_{m}$ is orthogonal and $U_{m}$ is upper-triangular. We form the rotation:
\begin{equation}
R_{m} =
\begin{cases}
O_{m}, & \det O_{m}=1\\[4pt]
\text{diag}(1,\dots,1,-1)\,O_{m}, & \det O_{m}=-1
\end{cases},
\end{equation}
where $R_{m}\in\mathrm{SO}(D)$.

The modality feature is then rotated:
\begin{equation}
\hat f^{m}=R_{m}^{\mathsf T}f^{m}.
\label{rotated}
\end{equation}

By \eqref{eq:isometry}, applying $R_m^{\mathsf T}$ preserves distances and inner products between features within the same representation. The mapping therefore performs modality-specific controlled adaptation without changing the internal geometry of that representation stream.

\paragraph{Lemma}
For any $R\in\mathrm{SO}(D)$ and $x,y\in\mathbb{R}^{D}$, \Cref{eq:isometry} holds.

\textit{Proof.} Let $R\in\mathrm{SO}(D)$ and $x,y\in\mathbb{R}^{D}$. We want to show that
\begin{equation}
\|Rx - Ry\|_{2} = \|x - y\|_{2}
\quad\text{and}\quad
\langle Rx, Ry\rangle = \langle x,y\rangle.
\end{equation}

First, consider the squared Euclidean norm of the difference:
\begin{equation}
\|Rx - Ry\|_{2}^{2}
= (Rx - Ry)^{\mathsf T}(Rx - Ry).
\end{equation}

Since matrix transpose distributes over addition and scalar multiplication,
\begin{equation}
(Rx - Ry)^{\mathsf T}
= x^{\mathsf T}R^{\mathsf T} - y^{\mathsf T}R^{\mathsf T},
\end{equation}
so that
\begin{equation}
\|Rx - Ry\|_{2}^{2}
= \bigl(x^{\mathsf T}R^{\mathsf T} - y^{\mathsf T}R^{\mathsf T}\bigr)
  (Rx - Ry).
\end{equation}

By associativity of matrix multiplication,
\begin{equation}
\|Rx - Ry\|_{2}^{2}
= x^{\mathsf T}R^{\mathsf T}R\,x
  - x^{\mathsf T}R^{\mathsf T}R\,y
  - y^{\mathsf T}R^{\mathsf T}R\,x
  + y^{\mathsf T}R^{\mathsf T}R\,y.
\end{equation}
Because $R$ is orthogonal, $R^{\mathsf T}R = I$. Substituting this identity gives
\begin{equation}
\|Rx - Ry\|_{2}^{2}
= x^{\mathsf T}x
  - x^{\mathsf T}y
  - y^{\mathsf T}x
  + y^{\mathsf T}y
= \|x - y\|_{2}^{2}.
\end{equation}

Taking square roots on both sides yields the first equality,
$\|Rx - Ry\|_{2} = \|x - y\|_{2}$.

Next, we verify preservation of the inner product. By definition,
\begin{equation}
\langle Rx, Ry\rangle
= (Rx)^{\mathsf T}(Ry)
= x^{\mathsf T}R^{\mathsf T}R\,y.
\end{equation}

Again using $R^{\mathsf T}R = I$, this reduces to
\begin{equation}
\langle Rx, Ry\rangle
= x^{\mathsf T}y
= \langle x,y\rangle.
\end{equation}

This completes the proof that applying the same element of $\mathrm{SO}(D)$ to two feature vectors preserves their Euclidean distance and inner product.

Since Triple Alignment is applied before orthogonal re-parameterisation, corresponding features from the three representations enter the transformation stage from an aligned latent space. Although the representation-specific transformations $R_m$ are learned independently, their outputs are jointly fused and optimized under the same downstream task objective. This allows each representation to retain its own geometry-preserving adaptation while coordinating its contribution to the final prediction. Triple Alignment thus provides explicit pre-fusion correspondence, while the fused task objective provides implicit task-driven coordination among the transformed representations.

\begin{figure*}[t]
    \centering
    \includegraphics[width=1\linewidth]{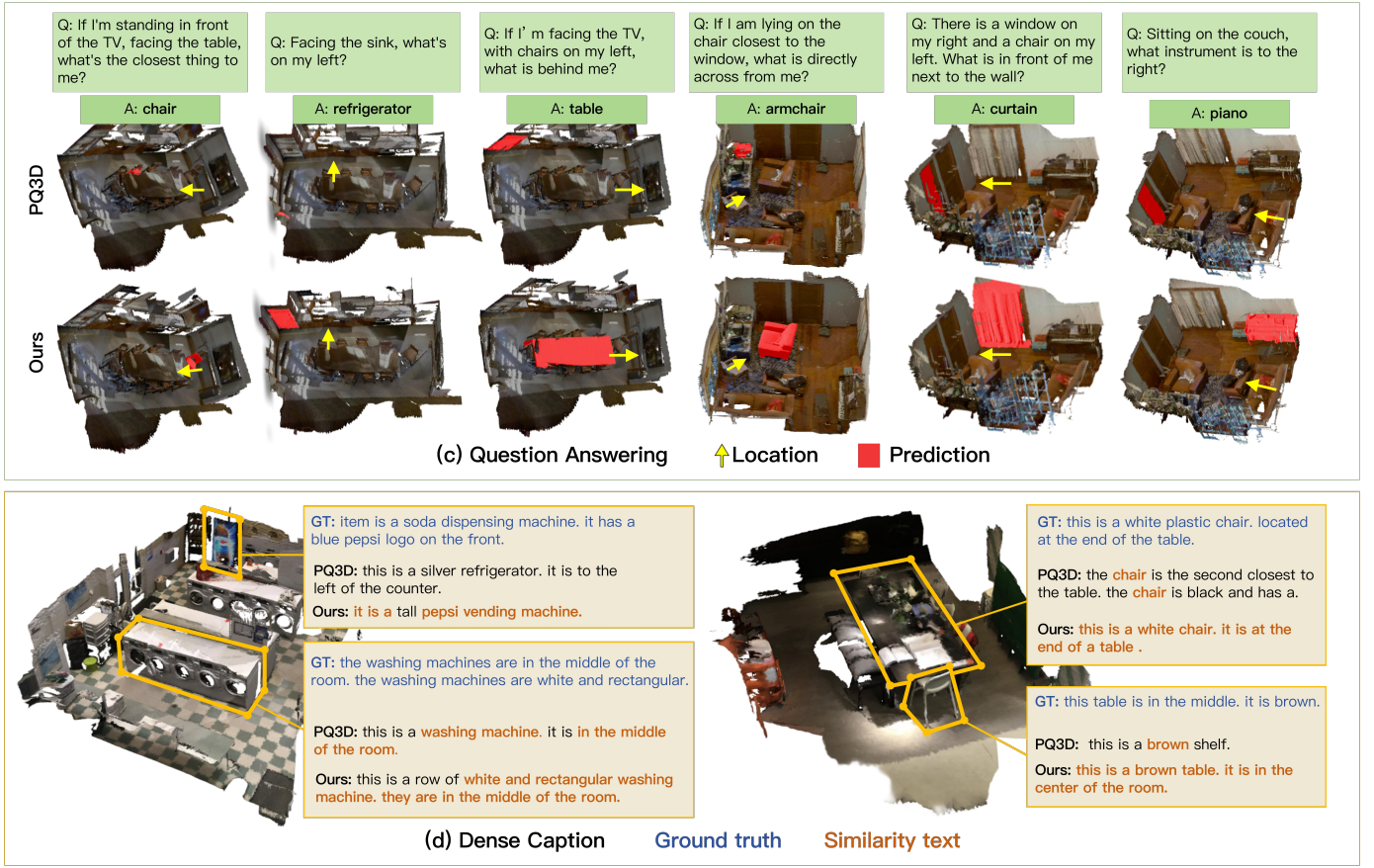}
    \caption{\textbf{Qualitative results on (c) question answering and (d) dense captioning.}
Representative predictions from our model are shown for different 3D scenes.}
    \label{fig_qua_2}
\end{figure*}

\paragraph{Adaptive fusion}
After orthogonal re-parameterisation, the three representation streams retain complementary scene information but need not contribute equally to the final prediction. A direct summation assigns equal importance to all representations, while an unconstrained weighted combination may alter the feature scale. We therefore employ a lightweight adaptive fusion mechanism to learn their relative contributions under downstream supervision. Specifically, we assign a learnable fusion weight to each of the three re-parameterized feature streams:
\begin{equation}
w=[w_{\mathcal I},w_{\mathcal P},w_{\mathcal V}]
\in\mathbb{R}^{3}.
\end{equation}
The corresponding fusion coefficients are obtained through softmax normalization:
\begin{equation}
\alpha_m
=
\frac{\exp(w_m)}
{\sum_{n\in\{\mathcal I,\mathcal P,\mathcal V\}}\exp(w_n)},
\qquad
m\in\{\mathcal I,\mathcal P,\mathcal V\},
\label{eq:fusion_weight}
\end{equation}
such that
\begin{equation}
\alpha_m\geq0,
\qquad
\sum_{m\in\{\mathcal I,\mathcal P,\mathcal V\}}\alpha_m=1.
\end{equation}
The transformed representation features are then fused as
\begin{equation}
\label{eq:fused}
f^{\mathrm{fused}}
=
\sum_{m\in\{\mathcal I,\mathcal P,\mathcal V\}}
\alpha_m\hat f^{m}.
\end{equation}

Because the fusion coefficients are non-negative and sum to one, \Cref{eq:fused} forms a convex combination of the transformed features. Together with the norm-preserving property of orthogonal re-parameterisation, the fused feature satisfies
\begin{equation}
\begin{aligned}
\|f^{\mathrm{fused}}\|_2
&\leq
\sum_m \alpha_m\|\hat f^m\|_2 \\
&=
\sum_m \alpha_m\|f^m\|_2 \\
&\leq
\max_m\|f^m\|_2.
\end{aligned}
\label{eq:fusion_bound}
\end{equation}
Thus, the fusion operation allows the relative contributions of the three representations to be learned without introducing unbounded feature-scale amplification through the fusion coefficients.

The fusion weights are optimized jointly with the representation-specific transformations and downstream prediction objectives. Here, ``adaptive'' refers to the learned adjustment of representation contributions during end-to-end training rather than input-dependent or query-dependent gating. The proposed modules are optimized jointly with the downstream task objectives. Specifically, the overall training objective is
\begin{equation}
\mathcal{L}_{\mathrm{total}}
=
\mathcal{L}_{\mathrm{task}}
+
\lambda_{\mathrm{align}}\mathcal{L}_{\mathrm{align}},
\end{equation}
where $\mathcal{L}_{\mathrm{task}}$ denotes the aggregate task-specific prediction loss and $\lambda_{\mathrm{align}}$ controls the contribution of Triple Alignment.

\begin{table}[t]
\caption{\textbf{Instance Segmentation results on the ScanNet200 validation set.} }
\label{tab:inst}
\centering
\small
\resizebox{\linewidth}{!}{\begin{tabular}{l|cccccc}
\toprule 
\multirow{2}{*}{Method}  & \multicolumn{6}{c}{ScanNet200} \\
 &  $\mathrm{AP}$ & $\mathrm{AP}_{50}$ & $\mathrm{AP}_{25}$ & head & common & tail \\
\midrule
Isbnet \cite{ngo2023isbnet}& 24.5&32.7&37.6&38.6&20.5&12.5\\
Mask3D \cite{schult2023mask3d}  & 26.9 & 36.2 & 41.4 & \underline{39.8} & 21.7 & 17.9  \\
PQ3D \cite{zhu2024unifying} & \underline{27.0} & \underline{38.9} & \underline{46.3} & 35.8 &\underline{24.2} & \underline{20.0} \\ \midrule
\rowcolor{LightGray}Ours  & \textbf{30.2} & \textbf{41.6}& \textbf{48.8} & \textbf{52.7}& \textbf{26.7}& \textbf{21.4}\\

\bottomrule
\end{tabular}}
\end{table}

\begin{table*}[t]
\caption{\textbf{Grounding performance (\%) on 3D visual grounding benchmarks.}
ScanRefer and Multi3DRefer are evaluated under an IoU threshold of 0.5, while Nr3D and Sr3D follow the standard evaluation protocol using ground-truth object masks. For Multi3DRefer, ZT, ST, and MT denote the zero-target, single-target, and multi-target settings, respectively.}
\label{tab:refer}
\centering
\resizebox{\linewidth}{!}{
\begin{tabular}{l|ccc|ccc|ccc|cccc}
\toprule
\multirow{2}{*}{Method} & \multicolumn{3}{c|}{ScanRefer} & \multicolumn{3}{c|}{Nr3D} & \multicolumn{3}{c|}{Sr3D} & \multicolumn{4}{c}{Multi3DRefer} \\
               & Unique & Multiple & Avg. & Easy & Hard & Avg. & Easy & Hard & Avg. & ZT & ST & MT & Avg. \\
\midrule
ViL3DRel \cite{vil3dref}       & 68.6 & 30.7 & 37.7 & 70.2 & 57.4 & 64.4 & 74.9 & 67.9 & 72.8 & -    & -    & -    & -    \\
3DJCG \cite{3djcg}          & 64.3 & 30.8 & 37.3 & -    & -    & -    & -    & -    & -    & \textbf{66.9} & 16.7 & 26.2 & 26.6 \\
UniT3D \cite{unit3d}         & 73.1 & 31.1 & 39.1 & -    & -    & -    & -    & -    & -    & -    & -    & -    & -    \\
M3DRef-CLIP \cite{multi3drefer}    & 77.2 & 36.8 & 44.7 & 55.6 & 43.4 & 49.4 & -    & -    & -    & 39.4 & 30.6 & 37.9 & 38.4 \\
3D-VisTA \cite{3d-vista}      & 75.1 & 39.1 & 45.8 & 72.1 & 56.7 & 64.2 & 78.8 & 71.3 & 76.4 & -    & -    & -    & -    \\
PQ3D \cite{zhu2024unifying}       & \underline{78.2} & \underline{46.2} & \underline{51.2} & \underline{75.0} & \underline{58.7} & \underline{66.7} & \underline{82.7} & \underline{72.8} & \underline{79.7} & 57.7 & \underline{43.6} & \underline{40.9} & \underline{50.1} \\
\midrule
\rowcolor{LightGray}Ours & \textbf{80.8} & \textbf{49.2} & \textbf{54.1} & \textbf{83.9} & \textbf{71.0} & \textbf{77.3} & \textbf{84.5} & \textbf{84.0} & \textbf{84.3} & \underline{59.3} & \textbf{48.2} & \textbf{46.5} & \textbf{54.2} \\
\bottomrule
\end{tabular}
}
\end{table*}

\begin{table*}[t]
\centering
\caption{\textbf{Answer accuracy on the ScanQA and SQA3D benchmarks.} For ScanQA, each entry reports the results on ``test with object'' and ``test without object'', respectively. In SQA3D, the results are reported using exact match (EM) accuracy between the predicted and ground-truth answers.}
\label{tab:scanqa_sqa3d}
\resizebox{\textwidth}{!}{\begin{tabular}{l|cccc|ccccccc}
\toprule
\multirow{2}{*}{Method} & \multicolumn{4}{c|}{ScanQA} & \multicolumn{7}{c}{SQA3D}  \\ 

& EM@1 & BLEU-1 & METEOR & CIDEr & What  & Is & How   & Can   & Which  & Other   & Avg.\\ 
\midrule
GPT-3 \cite{GTP3} & -&-&-&-&\textbf{39.7} & 46.0 & 40.5 & 45.6 & 36.1 & 38.4 & 41.0 \\ 
SQA3D \cite{sqa3d}& 23.5 / 20.9 & 31.6 / 30.7 & 13.6 / 12.6 & 67.3 / 60.2 & 31.6 & \underline{63.8} & \textbf{46.0} & \underline{69.5} & 43.9 & 45.3 & 46.6 \\
3D-VisTA \cite{3d-vista} & \underline{27.0} / \underline{23.0} & 34.4 / 30.2 & 15.2 / 12.9 & 76.6 / 62.6 & 34.8 & 63.3 & 45.4 & \textbf{69.8} & \textbf{47.2} & \underline{48.1} & \underline{48.5} \\  
PQ3D \cite{zhu2024unifying} & 26.1 / 20.0 & \underline{43.0} / \underline{36.1} & \underline{17.8} / \underline{13.9} & \underline{87.8} / \underline{65.2} & \underline{37.1} & 61.3 & 44.5 & 60.9 & 47.0 & 45.1 & 47.1 \\  \midrule 
\rowcolor{LightGray}Ours  & \textbf{28.2} / \textbf{23.5} & \textbf{46.4} / \textbf{38.0} & \textbf{19.6} / \textbf{16.5} & \textbf{89.4} / \textbf{67.3}& 36.5 & \textbf{63.9} & \underline{45.5} & 67.4 & \underline{47.1} & \textbf{50.8} & \textbf{48.7} \\
\bottomrule
\end{tabular}}
\end{table*}

\begin{table}[t]
\small
\centering
\caption{\textbf{Captioning performance on the Scan2Cap benchmark under the IoU@0.5 criterion}. }
\label{tab:scan2cap}
\resizebox{\linewidth}{!}{\begin{tabular}{l|cccc}
\toprule

\multirow{2}{*}{Method}  & \multicolumn{4}{c}{Scan2Cap} \\
 & CIDEr \quad & BLEU-4 \quad & METEOR \quad  & ROUGE \quad  \\ \midrule
Scan2Cap \cite{scan2cap} & 35.2 & 22.4 & 21.4 & 43.5 \\
3DJCG \cite{3djcg}   & 47.7 & 31.5 & 24.3  & 51.8 \\
3D-VisTA \cite{3d-vista} & 66.9 & 34.0 & 27.1 & 54.3 \\ 
PQ3D \cite{zhu2024unifying}& \underline{80.3} & \underline{36.0} & \underline{29.1} & \underline{57.9} \\ \midrule
\rowcolor{LightGray}Ours & \textbf{81.2} & \textbf{39.4} & \textbf{30.4} & \textbf{63.5} \\
\bottomrule
\end{tabular}}
\end{table}

\section{Experiment}

\subsection{Datasets, Metrics, and Implementation}
We evaluate models on eight public benchmarks spanning four task families. ScanNet200 \cite{scannet200} supplies instance masks, evaluated with AP, AP$_{50}$, AP$_{25}$, and head/common/tail category AP. ScanRefer \cite{scanrefer}, Nr3D \cite{referit3d}, Sr3D \cite{referit3d}, and Multi3DRefer \cite{multi3drefer} evaluate language-guided localization with their standard accuracy splits. ScanQA is evaluated with exact match, BLEU-1, METEOR, and CIDEr; SQA3D \cite{sqa3d} uses exact-match answer accuracy; and Scan2Cap  \cite{scan2cap} uses CIDEr, BLEU-4, METEOR, and ROUGE under IoU@0.5. We follow the benchmark protocols used by PQ3D \cite{zhu2024unifying}.

The training procedure consists of two stages. In the first stage, we train the model with instance segmentation alone on ScanNet200 for 800 epochs, using a classification head on instance queries instead of prompting object categories, and we guide the first 200 epochs with ground truth mask guidance \cite{fastinst} to speed convergence. In the second stage, we continue training on the full training set with all objectives for 50 epochs. We optimize with AdamW at a learning rate of $1\times10^{-4}$, batch size of 16, $\beta_{1}=0.9$, and $\beta_{2}=0.98$, applying a cosine decay learning rate scheduler. We follow the data augmentation strategy of PQ3D \cite{zhu2024unifying}, including geometric transformations and color augmentation during training. We set the number of queries to 120 to balance inference speed and the number of objects in the scene. All training is run on 4 NVIDIA A100 GPUs.

\subsection{Qualitative Results}

\Cref{fig_qua_1,fig_qua_2} present qualitative results across the four evaluated task families. In \Cref{fig_qua_1}, for instance segmentation, our model responds to both explicit category prompts, such as ``Printer'' and ``Drawer'', and more functional descriptions, such as ``I want to wash fruit'', demonstrating consistent language-conditioned scene understanding. In visual grounding, the advantage is more evident for expressions that require spatial disambiguation among nearby objects. For example, for ``the black trash can under the table'' and ``the first chair from the right'', our predictions align more closely with the ground-truth regions than those of PQ3D, suggesting better use of relational cues beyond object semantics. Similar behavior is observed in question answering in \Cref{fig_qua_2}, where egocentric questions involving directions such as left, behind, and across require reasoning with respect to the specified viewpoint; our predictions more consistently identify the corresponding answer-relevant objects. The dense-captioning examples further reveal differences in semantic and spatial understanding. PQ3D confuses the Pepsi vending machine with a refrigerator and the brown table with a shelf, whereas our model correctly identifies the object categories while retaining relevant attributes and spatial relations. For the washing-machine example, our description additionally captures its white, rectangular appearance and its location in the middle of the room. These examples qualitatively support the model's ability to associate language with object identity, attributes, and spatial structure across different 3D vision-language tasks. 

\begin{figure*}[t]
    \centering
    \includegraphics[width=1\textwidth]{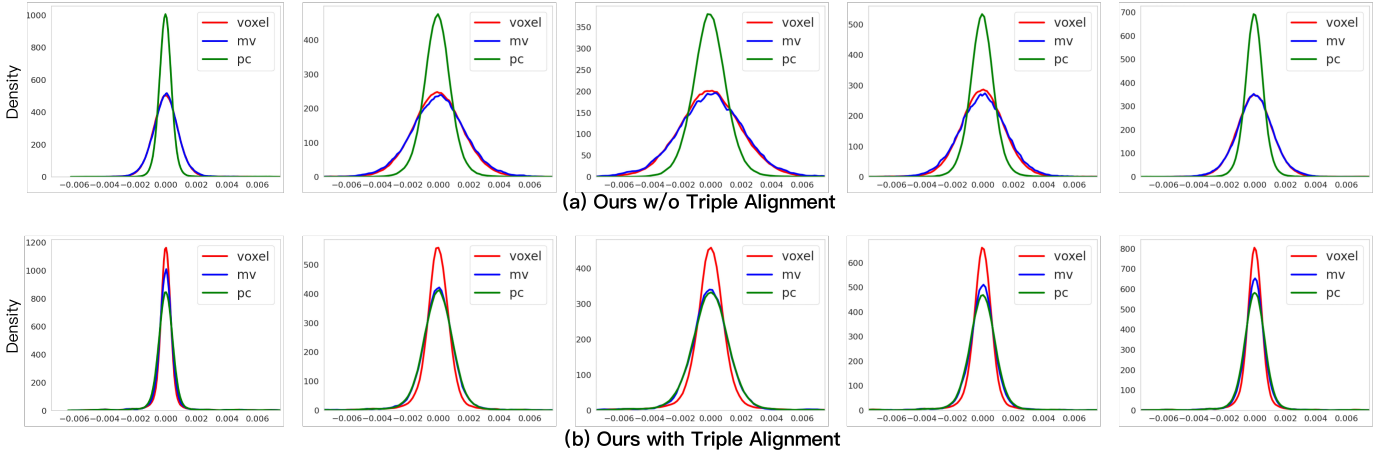}
   \caption{\textbf{Ablation study of Triple Alignment on feature distributions.} Density curves of voxel, multi-view (MV), and point-cloud (PC) features are shown for representative ScanNet200 scenes, comparing (a) our model without Triple Alignment and (b) the full model. Triple Alignment leads to more consistent feature distributions across the three representations.}

    \label{ala_1}
\end{figure*}

\begin{table}[t]
\caption{\textbf{Ablation Study: Triple Alignment (TA) and Orthogonal Fusion (OF) for ScanNet200 Instance Segmentation.}}
\label{tab:ablation_is}
\centering
\begin{tabular}{cc|ccc}
\toprule
TA & OF & AP & AP$_{50}$ & AP$_{25}$ \\
\midrule
- & - & 27.0 & 38.9 & 46.3 \\
\checkmark & - & \underline{30.0} & 41.1 & 47.2 \\
- & \checkmark & 29.8 & \underline{41.3} & \underline{47.6} \\
\rowcolor{LightGray}\checkmark & \checkmark & \textbf{30.2} & \textbf{41.6} & \textbf{48.8} \\
\bottomrule
\end{tabular}
\end{table}

\begin{table}[t]
\caption{\textbf{Ablation Study: Triple Alignment (TA) and Orthogonal Fusion (OF) for Visual grounding.} ``TA'' is short for Triple Alignment, ``OF'' is short for Orthogonal Fusion.}
\label{tab:ablation_vg}
\centering
\resizebox{\linewidth}{!}{
\begin{tabular}{cc|cccc}
\toprule
TA & OF & ScanRefer & Nr3D & Sr3D & Multi3DRefer \\
\midrule
- & - & 51.2 & 66.7 & 79.7 & 50.1 \\
\checkmark & - & \underline{53.3} & 71.8 & 80.5 & \underline{53.1} \\
- & \checkmark & 52.8  & \underline{75.4} & \underline{83.7} & 51.9 \\
\rowcolor{LightGray}\checkmark & \checkmark & \textbf{54.1}  & \textbf{77.3} & \textbf{84.3} & \textbf{54.2} \\
\bottomrule
\end{tabular}
}
\end{table}

\begin{table}[t]
\caption{\textbf{Ablation Study: Triple Alignment (TA) and Orthogonal Fusion (OF) for Question Answering on ScanQA.}}
\label{tab:ablation_qa}
\centering
\begin{tabular}{cc|cccc}
\toprule
TA & OF & EM@1 & BLEU-1 & METEOR & CIDEr \\
\midrule
- & - & 26.1 / 20.0 & 43.0 / 36.1 & 17.8 / 13.9 & \underline{87.8} / 65.2 \\
\checkmark & - & 27.5 / 21.4 & \underline{45.8} / 38.0 & 19.2 / 15.2 & 89.0 / 66.8 \\
- & \checkmark & \underline{27.9} / \underline{21.8} & \textbf{46.4} / \textbf{38.5} & \underline{18.8} / \underline{14.5} & 88.5 / \underline{67.1} \\
\rowcolor{LightGray}\checkmark & \checkmark & \textbf{28.2} / \textbf{23.5} & \textbf{46.4} / 38.0 & \textbf{19.6} / \textbf{16.5} & \textbf{89.4} / \textbf{67.3} \\
\bottomrule
\end{tabular}
\end{table}

\begin{table}[t]
\caption{\textbf{Ablation Study: Triple Alignment (TA) and Orthogonal Fusion (OF) for Answer Accuracy on SQA3D.}}
\label{tab:ablation_sqa3d}
\centering
\begin{tabular}{cc|ccccccc}
\toprule
TA & OF & What & Is & How & Can & Which & Other & Avg. \\
\midrule
- & - & \underline{37.1} & 61.3 & 44.5 & 60.9 & 47.0 & 45.1 & 47.1  \\
\checkmark & - & 36.6 & 62.1 & \underline{45.2} & \underline{63.7} & 46.1 & 46.2 & \textbf{50.8} \\
- & \checkmark & \textbf{37.5} & \underline{62.3} & 44.8 & 62.0 & \textbf{47.2} & \underline{49.5} & \underline{49.9} \\
\rowcolor{LightGray}\checkmark & \checkmark & 36.5 & \textbf{63.9} & \textbf{45.5} & \textbf{67.4} & \underline{47.1} & \textbf{50.8} & 48.7  \\
\bottomrule
\end{tabular}
\end{table}

\begin{table}[t]
\caption{\textbf{Ablation Study: Triple Alignment (TA) and Orthogonal Fusion (OF) for Caption Performance on Scan2Cap.}}
\label{tab:ablation_cap}
\centering
\begin{tabular}{cc|cccc}
\toprule
TA & OF & CIDEr & BLEU-4 & METEOR & ROUGE \\
\midrule
– & – & 80.3 & 36.0 & 29.1 & 57.9 \\
\checkmark & – & \textbf{81.3} & 38.5 & \textbf{30.7} & 62.4 \\
– & \checkmark & 80.7 & \underline{38.9} & 29.8 & \underline{62.5} \\
\rowcolor{LightGray}\checkmark & \checkmark & \underline{81.2} & \textbf{39.4} & \underline{30.4} & \textbf{63.5} \\
\bottomrule
\end{tabular}
\end{table}

\subsection{Quantitative Results}

\paragraph{Instance Segmentation} As shown in \Cref{tab:inst}, our method achieves the best performance among the compared methods on ScanNet200 across all evaluation metrics. Compared with representative methods such as ISBNet \cite{ngo2023isbnet}, Mask3D \cite{schult2023mask3d}, and PQ3D \cite{zhu2024unifying}, our approach improves the overall Average Precision (AP) from 27.0 to 30.2, and further increases $\mathrm{AP}_{50}$ and $\mathrm{AP}_{25}$ from 38.9 and 46.3 to 41.6 and 48.8, respectively. In terms of long-tailed class performance, our method substantially improves the head-class AP to 52.7, surpassing PQ3D by 16.9 points, while also improving the common and tail categories from 24.2 and 20.0 to 26.7 and 21.4.

Unlike approaches that directly operate on individual or heterogeneous representations, our method explicitly establishes cross-representation correspondence before their joint use. Triple Alignment first aligns spatially corresponding features, after which representation-specific orthogonal re-parameterisation enables controlled adaptation while preserving intra-representation geometry, followed by adaptive fusion under downstream supervision. The consistent improvements across head, common, and tail categories demonstrate the effectiveness of integrating complementary representations within this align-then-fuse framework.

\begin{figure*}[t]
    \centering
    \includegraphics[width=1\textwidth]{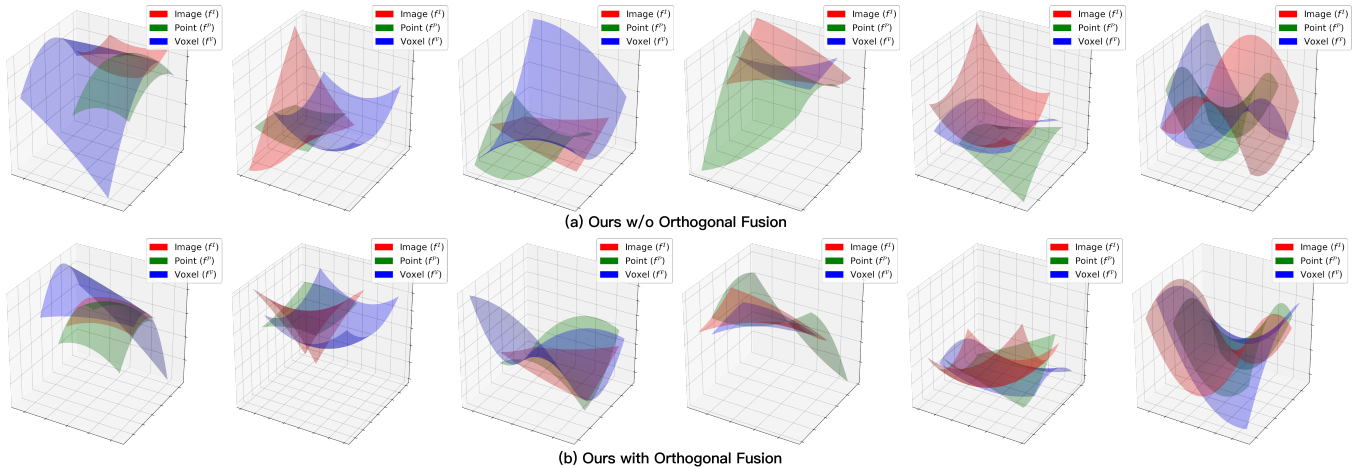}
    \caption{\textbf{Ablation study on Orthogonal Fusion.} For representative examples, we fit bivariate polynomial surfaces to the feature distributions of the image, point-cloud, and voxel representations, comparing the variant without Orthogonal Fusion (top) and the full model (bottom). After orthogonal re-parameterisation with \Cref{rotated}, the fitted surfaces exhibit greater overlap, indicating more consistent coordination among the three representations before fusion.}
    \label{ala_2}
\end{figure*}
\paragraph{Visual Grounding} As shown in \Cref{tab:refer}, we evaluate grounding accuracy on four benchmark datasets, including ScanRefer, Nr3D, Sr3D, and Multi3DRefer. Compared with PQ3D, our method improves the average accuracy by 2.9, 10.6, and 4.6 points on ScanRefer, Nr3D, and Sr3D, respectively. On Multi3DRefer, the average accuracy increases from 50.1 to 54.2, with the ST and MT scores improving from 43.6 and 40.9 to 48.2 and 46.5, respectively. For the ZT setting, our method achieves 59.3, improving upon PQ3D's 57.7 but remaining below the best result of 66.9 from 3DJCG. Overall, the consistent gains across the four grounding benchmarks indicate that explicitly aligning heterogeneous scene representations before their controlled adaptation and fusion is particularly beneficial for language-conditioned object localization.

\paragraph{Question Answering} As shown in \Cref{tab:scanqa_sqa3d}, our method achieves the best overall results on ScanQA across both exact-match and language-generation metrics. On the `test with object'' split, our model improves EM@1 from the previous best of 27.0 to 28.2, while increasing BLEU-1, METEOR, and CIDEr over PQ3D from 43.0, 17.8, and 87.8 to 46.4, 19.6, and 89.4, respectively. On the `test without object'' split, EM@1 increases from 23.0 to 23.5, while BLEU-1, METEOR, and CIDEr improve from 36.1, 13.9, and 65.2 to 38.0, 16.5, and 67.3. These improvements show that the benefit of combining aligned complementary representations extends beyond object localization to language-conditioned scene reasoning.

On SQA3D, our method achieves the highest average accuracy of 48.7, slightly exceeding the previous best result of 48.5. It also obtains the best performance for the `Is'' and `Other'' question categories, reaching 63.9 and 50.8, respectively, while remaining competitive across the other question types. Although the overall improvement is modest compared with that observed on the grounding benchmarks, the results indicate that the proposed representation integration strategy remains effective for situated question answering without uniformly improving every question category.

\paragraph{Dense Captioning} As shown in \Cref{tab:scan2cap}, our method achieves the best performance among the compared approaches across all four Scan2Cap metrics. Compared with PQ3D, CIDEr increases from 80.3 to 81.2, BLEU-4 from 36.0 to 39.4, METEOR from 29.1 to 30.4, and ROUGE from 57.9 to 63.5. In particular, the larger gains on BLEU-4 and ROUGE indicate improved agreement between the generated descriptions and reference captions, while the improvements in CIDEr and METEOR further confirm the overall captioning quality. Together with the results on segmentation, grounding, and question answering, these results demonstrate that the proposed align-then-fuse formulation generalizes across both perception-oriented and language-generation tasks.

\begin{table}[t]
\caption{\textbf{Ablation Study: Fusion methods for Instance Segmentation on ScanNet200.}}
\label{tab:ablation_is_fm}
\centering
\begin{tabular}{c|ccc}
\toprule
Methods & AP & AP$_{50}$ & AP$_{25}$ \\
\midrule
\rowcolor{LightGray}AF & \textbf{30.2} & \textbf{41.6} & \textbf{48.8} \\
$\mathcal I \oplus \mathcal P \oplus \mathcal V$ & 28.2 & 39.2 & 45.4 \\
$\mathcal I \otimes \mathcal P \otimes \mathcal V$ & \underline{30.0} & \underline{41.2} & \underline{48.3} \\
$\mathcal I \oplus \mathcal P \otimes \mathcal V$ & 28.5 & 40.6 & 47.1 \\
$\mathcal I \otimes \mathcal P \oplus \mathcal V$ & 28.7 & 40.5 & 47.3 \\
\bottomrule
\end{tabular}
\end{table}

\begin{table}[t]
\caption{\textbf{Ablation Study: Fusion methods for Visual grounding.} ``AF'' represents Adaptive fusion, $\oplus$ represents element-by-element addition. $\otimes$ represents feature concatenation and is mapped through a fully connected layer.}
\label{tab:ablation_vg_fm}
\centering
\begin{tabular}{c|cccc}
\toprule
Methods & ScanRefer & Nr3D & Sr3D & Multi3DRefer \\
\midrule
\rowcolor{LightGray}AF & \textbf{54.1} & \textbf{77.3} & \textbf{84.3} & \underline{54.2} \\
$\mathcal I \oplus \mathcal P \oplus \mathcal V$ & 49.3 & 65.5 & 78.3 & 44.4 \\
$\mathcal I \otimes \mathcal P \otimes \mathcal V$ & \underline{53.6} & \underline{75.2} & \underline{81.9} & \textbf{55.1} \\
$\mathcal I \oplus \mathcal P \otimes \mathcal V$ & 50.2 & 72.1 & 79.7 & 48.0 \\
$\mathcal I \otimes \mathcal P \oplus \mathcal V$ & 50.4 & 71.8 & 80.2 & 47.6 \\
\bottomrule
\end{tabular}
\end{table}

\begin{table}[t]
\caption{\textbf{Ablation Study: Fusion Methods for Answer Accuracy on ScanQA.}}
\label{tab:ablation_qa_fm}
\centering
\begin{tabular}{c|cccc}
\toprule
Methods & EM@1 & BLEU-1 & METEOR & CIDEr \\
\midrule
\rowcolor{LightGray}AF & \textbf{28.2 / 23.5} & \textbf{46.4 / 38.0} & \textbf{19.6 / 16.5} & \textbf{89.4 / 67.3} \\
$\mathcal I \oplus \mathcal P \oplus \mathcal V$ & 25.0 / 19.9 & 42.2 / 34.1 & 17.8 / 13.9 & 82.0 / 60.8 \\
$\mathcal I \otimes \mathcal P \otimes \mathcal V$ & \underline{28.0 }/ \textbf{23.5} & \underline{45.8 / 37.5} & \underline{19.2 / 16.0} & \underline{89.2} / \textbf{67.3} \\
$\mathcal I \oplus \mathcal P \otimes \mathcal V$ & 26.3 / 21.5 & 43.4 / 36.3 & 18.0 / 14.8 & 86.4 / 64.2 \\
$\mathcal I \otimes \mathcal P \oplus \mathcal V$ & 26.8 / \underline{21.8} & 44.0 / 36.5 & 18.3 / 15.2 & 87.5 / \underline{65.1} \\
\bottomrule
\end{tabular}
\end{table}

\begin{table}[t]
\caption{\textbf{Ablation Study: Fusion Methods for Answer Accuracy on SQA3D.}}
\label{tab:ablation_sqa3d_fm}
\centering
\begin{tabular}{c|ccccccc}
\toprule
Methods & What & Is & How & Can & Which & Other & Avg. \\
\midrule
\rowcolor{LightGray}AF                               & \textbf{36.5} & \textbf{63.9} & \textbf{45.5} & \textbf{67.4} & \textbf{47.1} & \textbf{50.8} & \textbf{48.7} \\
$\mathcal I \oplus \mathcal P \oplus \mathcal V$ & 34.7 & 61.2 & 43.9 & 60.3 & 45.5 & 46.8 & 42.3 \\
$\mathcal I \otimes \mathcal P \otimes \mathcal V$ & \underline{36.2} & \textbf{63.9} & \textbf{45.5} & \textbf{67.4} & \underline{47.2} & \textbf{50.8} & \underline{48.2} \\
$\mathcal I \oplus \mathcal P \otimes \mathcal V$ & 35.6 & \underline{62.5} & \underline{44.5} & 62.4 & 46.7 & 48.2 & 46.5 \\
$\mathcal I \otimes \mathcal P \oplus \mathcal V$ & 35.5 & 62.2 & 44.2 & \underline{63.8} & 46.5 & \underline{49.3} & 47.4 \\
\bottomrule
\end{tabular}
\end{table}

\begin{table}[t]
\caption{\textbf{Ablation Study: Fusion Methods for captioning performance on Scan2Cap.}}
\label{tab:ablation_cap_fm}
\centering
\begin{tabular}{c|cccc}
\toprule
Methods & CIDEr & BLEU-4 & METEOR & ROUGE \\
\midrule
\rowcolor{LightGray}AF & \textbf{81.2} & \textbf{39.4} & \textbf{30.4} & \textbf{63.5} \\
$\mathcal I \oplus \mathcal P \oplus \mathcal V$ & 78.3 & 35.7 & 28.1 & 55.9 \\
$\mathcal I \otimes \mathcal P \otimes \mathcal V$ & \underline{81.6} & \underline{39.0} & \underline{30.2} & \underline{63.8} \\
$\mathcal I \oplus \mathcal P \otimes \mathcal V$ & 79.5 & 37.5 & 29.5 & 60.4 \\
$\mathcal I \otimes \mathcal P \oplus \mathcal V$ & 79.8 & 38.5 & 29.7 & 61.3 \\
\bottomrule
\end{tabular}
\end{table}

\subsection{Ablation Study}

\paragraph{Ablation on Triple Alignment and Orthogonal Fusion}
\Cref{ala_1} provides an illustration of the effect of introducing the Triple Alignment module on the feature distributions across different representations. In the original setting, observable discrepancies exist among the representations, as reflected by differences in peak position and distribution width. After introducing Triple Alignment, the density curves become more consistent in both peak location and distribution width, indicating reduced representation discrepancy in the latent space. This observation is consistent with the role of Triple Alignment in establishing correspondence among spatially corresponding scene features. As shown in \Cref{tab:ablation_vg}, introducing Triple Alignment alone improves the visual grounding accuracy over the baseline by 2.1, 5.1, 0.8, and 3.0 points on ScanRefer, Nr3D, Sr3D, and Multi3DRefer, respectively. When Orthogonal Fusion is already enabled, further introducing Triple Alignment still yields improvements of 1.3, 1.9, 0.6, and 2.3 points on the four benchmarks. These results demonstrate that explicitly establishing segment-level correspondence before fusion consistently benefits language-conditioned object localization.

\Cref{ala_2} qualitatively illustrates the effect of Orthogonal Fusion on multi-representation feature coordination. We fit bivariate polynomial surfaces to features extracted from the image, point-cloud, and voxel representations for representative examples, comparing the variant without Orthogonal Fusion with the full model. After orthogonal re-parameterisation, the fitted surfaces generally exhibit greater overlap, providing a qualitative indication that the representation-specific features are more consistently coordinated before fusion. Importantly, this effect does not imply that the three representations are forced into an identical feature distribution. Instead, the orthogonal transformations provide representation-specific adaptation while preserving pairwise distances and inner products within each representation.

The quantitative results further support this observation. As shown in \Cref{tab:ablation_is}, introducing Orthogonal Fusion alone improves AP, AP$_{50}$, and AP$_{25}$ on ScanNet200 by 2.8, 2.4, and 1.3 points, respectively. Similar improvements are observed for visual grounding: compared with the baseline, Orthogonal Fusion alone improves performance by 1.6, 8.7, 4.0, and 1.8 points on ScanRefer, Nr3D, Sr3D, and Multi3DRefer, respectively. On ScanQA (\Cref{tab:ablation_qa}), it improves BLEU-1 by 3.4/2.4 points and CIDEr by 0.7/1.9 points on the test sets with/without object annotations. The results on SQA3D (\Cref{tab:ablation_sqa3d}) and Scan2Cap (\Cref{tab:ablation_cap}) further show that both components provide complementary benefits across question answering and captioning, although neither consistently dominates every individual metric. Together, these results suggest that geometry-preserving re-parameterisation provides effective representation-specific adaptation before cross-representation fusion.

\paragraph{Ablation for Fusion Methods}

As shown in \Cref{tab:ablation_is_fm} to~\ref{tab:ablation_cap_fm}, we systematically compare our adaptive fusion with alternative multi-representation fusion strategies based on element-wise addition ($\oplus$) and feature concatenation followed by projection ($\otimes$).

Our Adaptive Fusion achieves the strongest overall performance across the evaluated tasks. It obtains the best results on all three ScanNet200 metrics and on three of the four visual-grounding benchmarks, while full concatenation performs best on Multi3DRefer with 55.1. A similar pattern is observed for the language tasks: Adaptive Fusion achieves the best or tied-best results across the ScanQA metrics and the highest average accuracy of 48.7 on SQA3D. On Scan2Cap, it achieves the best BLEU-4 and METEOR scores of 39.4 and 30.4, whereas full concatenation obtains slightly higher CIDEr and ROUGE scores of 81.6 and 63.8.

Overall, full concatenation generally performs better than direct element-wise addition, suggesting that simply combining the representations with equal contribution is insufficient. In comparison, our Adaptive Fusion assigns normalized learnable coefficients to the transformed representations and optimizes their relative contributions jointly under downstream task supervision. Its consistently strong performance across different task families supports the use of learned representation coordination after alignment and orthogonal re-parameterisation, while the few metrics favoring concatenation also indicate that no single fusion rule uniformly dominates every evaluation setting.

\begin{table}[t]
\centering
\caption{\textbf{Model complexity comparison under prompt-conditioned inference.}}
\label{tab:model_complexity}
\small
\resizebox{\linewidth}{!}{\begin{tabular}{lccc}
\toprule
\textbf{Model} & \textbf{Parameters} & \textbf{Single Forward (s)} & \textbf{Forward FLOPs} \\
\midrule
PQ3D & \textbf{247,875,873 (247.88M)} & \textbf{0.1340} & \textbf{25,808,556,116 (25.81 GFLOPs)} \\
\rowcolor{LightGray} Ours & \underline{254,953,773 (254.95M)} & \underline{0.1648} & \underline{26,941,755,488 (26.94 GFLOPs)} \\
\bottomrule
\end{tabular}}
\end{table}

\subsection{Model Complexity Comparison.}

We further compare the computational complexity of our method with PQ3D under prompt-conditioned inference. As shown in Table~\ref{tab:model_complexity}, PQ3D contains 247.88M parameters and requires 25.81 GFLOPs for a single forward pass, whereas our model contains 254.95M parameters and requires 26.94 GFLOPs. This corresponds to an increase of approximately 7.08M parameters (2.9\%) and 1.13 GFLOPs (4.4\%), respectively. The inference time increases from 0.1340~s to 0.1648~s per forward pass, corresponding to an additional 0.0308~s (23.0\%). The additional inference cost mainly arises from the representation-specific orthogonal re-parameterisation and Adaptive Fusion operations. Overall, our method introduces a modest increase in model complexity while consistently improving performance across the evaluated 3D vision-language tasks.

\section{Discussion and Limitations}

The results support a specific proposition: alignment and fusion solve different parts of heterogeneous representation integration. Triple Alignment improves cross-representation correspondence before task decoding, whereas Orthogonal Fusion restricts the subsequent re-parameterisation so that it does not alter pairwise distances or inner products within each representation. Their combination is consistently strongest in the visual-grounding ablation, although the ablations across the remaining tasks show that neither component dominates every individual metric. These findings should be interpreted within the tested indoor-scene benchmarks and pretrained feature pipelines.

The method relies on pretrained, representation-specific encoders and therefore does not establish fully end-to-end cross-representation learning. Orthogonal re-parameterisation and Adaptive Fusion introduce additional inference cost: relative to PQ3D, the measured model has 254.95M rather than 247.88M parameters, 26.94 rather than 25.81 GFLOPs, and a 0.1648-s rather than 0.1340-s forward pass. Evaluation is limited to the eight reported datasets, which largely share indoor-scene structure; robustness to outdoor scenes, missing representations, and sensor corruption remains untested. These limitations motivate more efficient representation adaptation and fusion, as well as explicit evaluation under missing-representation settings.

\section{Conclusion}
We present an ``\textit{align-then-fuse}" framework for 3D vision--language understanding that integrates point clouds, voxel grids, and multi-view images within a unified architecture. Our Triple Alignment establishes segment-level correspondence across heterogeneous representations before task decoding, while representation-specific orthogonal re-parameterisation enables controlled adaptation by preserving the pairwise geometry within each representation. The resulting features are then combined through Adaptive Fusion under downstream task supervision. Extensive experiments on eight datasets show that our model achieves competitive or state-of-the-art performance across tasks ranging from low-level segmentation to high-level reasoning.

\bibliographystyle{IEEEtran}
\bibliography{references}
\end{document}